\documentclass[sigconf]{acmart}
\usepackage{multirow}
\usepackage{graphicx} 
\usepackage{CJKutf8}
\usepackage{tikz} 

\AtBeginDocument{%
  }

\setcopyright{none}
\renewcommand\footnotetextcopyrightpermission[1]{}
\acmConference[PoliSim@CHI 2026]{PoliSim@CHI 2026: LLM Agent Simulation for Policy}{April 16, 2026}{Barcelona, Spain}
\renewcommand{\footnotetextcopyrightpermission}[1]{%
  \footnotetext{%
    This paper was prepared for \href{https://polisim.net/}{PoliSim@CHI 2026: LLM Agent Simulation for Policy}, a workshop at \href{https://chi2026.acm.org/}{CHI 2026 (CHI Conference on Human Factors in Computing Systems)}, April 16, 2026, Barcelona, Spain.
  }%
}

\begin{document}
\definecolor{mainblue}{HTML}{D3E1F1}
\newcommand{\circnum}[1]{%
  \tikz[baseline=(char.base)]{
    \node[shape=circle, draw, fill=mainblue, inner sep=0.93pt] (char) {#1};
  }%
}
\title{Cross-Cultural Simulation of Citizen Emotional Responses\\ to Bureaucratic Red Tape Using LLM Agents}

\author{Wanchun Ni}
\affiliation{%
  \institution{ETH Zurich}
  \city{Zurich}
  \country{Switzerland}
}
\email{wanchun.ni@inf.ethz.ch}

\author{Jiugeng Sun}
\affiliation{%
  \institution{ETH Zurich}
  \city{Zurich}
  \country{Switzerland}
}
\email{jiugeng.sun@inf.ethz.ch}

\author{Yixian Liu}
\affiliation{%
  \institution{City University of Hong Kong}
  \city{Hong Kong SAR}
\country{China}}
\email{yixianliu9-c@my.cityu.edu.hk}

\author{Mennatallah El-Assady}
\affiliation{%
  \institution{ETH Zurich}
  \city{Zurich}
  \country{Switzerland}
}
\email{menna.elassady@ai.ethz.ch}

\renewcommand{\shortauthors}{Ni et al.}

\begin{abstract}

  Improving policymaking is a central concern in public administration. Prior human subject studies reveal substantial cross-cultural differences in citizens' emotional responses to red tape during policy implementation. While LLM agents offer opportunities to simulate human-like responses and reduce experimental costs, their ability to generate culturally appropriate emotional responses to red tape remains unverified. To address this gap, we propose an evaluation framework for assessing LLMs' emotional responses to red tape across diverse cultural contexts.As a pilot study, we apply this framework to a single red-tape scenario. Our results show that all models exhibit limited alignment with human emotional responses, with notably weaker performance in Eastern cultures. Cultural prompting strategies prove largely ineffective in improving alignment. We further introduce \textbf{RAMO}, an interactive interface for simulating citizens' emotional responses to red tape and for collecting human data to improve models. The interface is publicly available at \url{https://ramo-chi.ivia.ch}.

\end{abstract}

\ccsdesc[500]{Computing methodologies~Simulation evaluation}

\keywords{LLM Simulation, LLM-Agents, Social Computing, Social Simulation, LLM Evaluation}


\maketitle

\section{Introduction}

Effective policymaking requires understanding citizens' emotional responses to administrative procedures. Complicated policy procedures can contribute to citizens' emotional distress. Bureaucratic red tape (hereafter \textit{red tape}) refers to unnecessary procedural overhead in policies that impose costs without delivering functional value. As a pathological feature of governance, red tape is characterised by dysfunctional rules that remain in force without serving legitimate purposes. It becomes problematic when compliance burdens outweigh the benefits, turning administrative processes into systemic barriers rather than effective governance mechanisms. Existing studies of citizens' emotional responses to red tape primarily rely on controlled human experiments. To avoid bias arising from prior exposure or learning effects, these experiments typically require recruiting new participants for each study, making the process time-consuming, costly, and inherently limited in scale and coverage.

Large Language Models (LLMs) have opened new possibilities for simulating human behaviour \cite{park2023generative}, offering a cost-effective means to model public responses at scale.
Such simulations can serve as testbeds for evaluating and refining policy designs prior to real-world deployment. Recent studies further explore LLM-based social simulations by assigning agents diverse personas, enabling the modelling of complex societal dynamics while substantially reducing the cost and logistical burden of traditional human-subject experiments. However, despite their impressive linguistic and world knowledge, LLMs are known to embed cultural biases that can distort simulated social responses. Prior work has shown that LLMs exhibit systematic biases in cultural contexts, with their outputs often reflecting Western-centric perspectives \cite{tao2024cultural}. Moreover, emotional responses may exhibit different patterns of bias compared to factual or attitudinal judgments, yet remain underexplored in existing studies. These limitations raise a critical question for the facilitation of LLM-based agents in real-world policymaking: \textbf{To what extent can LLM-based agents reproduce human-like, culture-specific emotional responses to bureaucratic red tape?}

In this study, we examine LLM agents' emotional responses to red tape across three representative cultural contexts on one scenario as a pilot study: Western (Germany), hybrid Western--Eastern (Hong Kong SAR), and Eastern (Mainland China). We compare model-generated emotions with human experimental results to evaluate emotional alignment and inherent cultural biases. Given the limited availability of complete human datasets, we assess alignment using top-3 emotion overlap and between-group statistical significance. Our results show that most LLMs exhibit limited alignment with human emotional responses, and their performance degrades substantially in Eastern cultural contexts. These findings suggest that simple prompt-based cultural conditioning is insufficient to override models' inherent cultural biases. More sophisticated mechanisms are therefore required before LLM-agents can be reliably deployed to support real-world policymaking. Nevertheless, GPT-5 demonstrates promising capabilities in capturing cultural nuances through in-context learning from culturally grounded prompts. Furthermore, we present \textbf{RAMO} (\textbf{\underline{R}}ed T\textbf{\underline{a}}pe E\textbf{\underline{m}}otional Simulat\textbf{\underline{o}}r), a policymaker-oriented interface for simulating citizens' emotional responses under potential red tape. RAMO aims to optimise policymaking by revealing how citizens' emotions shift across different red-tape scenarios and by enabling the collection of self-reported emotional data for future development.

\section{Related Works}
\textbf{Red Tape} Red tape is defined as \textit{rules, regulations, and procedures that remain in force and entail a compliance burden for designated stakeholders but whose contribution to stakeholders' objectives or values is less than the compliance and implementation resources expended on the rule}~\cite{rtdefi}. To examine how citizens perceive red tape and the extent to which it is experienced as burdensome, traditional research has mostly relied on self-reported survey experiments \cite{rtsv1, rtsv2, rtsv3}. More recent laboratory studies introduce objective biological measures to capture micro-level affective responses to red tape \cite{hattke2020emotional, liu2026redtape}. However, these empirical methods are time-consuming and labor-intensive, and constrained by limited sample sizes and cultural coverage. As a complementary approach, scalable and cost-effective LLM-driven agents can simulate citizens' emotional responses to red tape, thereby supporting policymakers in improving policymaking processes.

\textbf{LLM-Agent Social Simulation} Large Language Models (LLMs) have enabled new approaches to simulate human behaviour in social contexts, with recent frameworks spanning from economics~\cite{ji-etal-2024-srap, li2024econagentlargelanguagemodelempowered}, political surveys~\cite{chuang-etal-2024-simulating, jiang2024donald}, cross-cultural analysis~\cite{ki-etal-2025-multiple}, to other broad social domains~\cite{bougie-watanabe-2025-citysim, zhang-etal-2025-trendsim, tang2025gensimgeneralsocialsimulation, wang2025decodingechochambersllmpowered, wang-etal-2024-towards-objectively}. However, these simulations exhibit instability and systematic biases \cite{chuang-etal-2024-beyond, Taubenfeld24, Qi25}, including limited cultural knowledge~\cite{wu-etal-2025-socialcc}, sensitivity to prompting \cite{Bisbee2024Synthetic}, and tendencies to overestimate positive feedback in opinion surveys~\cite{10.1145/3708319.3733685}. To address these challenges, recent work has improved alignment through multi-step reasoning~\cite{yu2025largescalesimulationlargelanguage}, the incorporation of demographic and issue-related covariates~\cite{Lee_2024}, agent-based frameworks~\cite{Li25}, and conditioning on individuals' past opinions~\cite{hwang-etal-2023-aligning}, with evidence that LLMs can replicate certain cultural differences when appropriately configured~\cite{jiang2024donald}.

\section{Simulation}
To address the research question: \textbf{Can LLMs simulate culture-aware, human-aligned emotional responses to red tape?} We develop a cross-cultural simulation framework that integrates diverse personas and country-level cultural factors into dynamic red tape scenarios. Our alignment assessment comprises two steps: replicating human experiments across three cultures and conducting model probing analyses.
\subsection{Human Experiment on Red Tape}
\label{subsec: gt}

We use results from three prior human experiments as ground truth. The Germany data come from \citet{hattke2020emotional}, who pioneered neurological methods in public administration through a lab experiment using facial expression analysis. Participants encountered simulated government red tape scenarios under two conditions: delay (time-wasting) and burden (extra paperwork). Facial expressions were captured via webcam and analysed to infer emotional responses. \citet{liu2026redtape} replicated this study in Hong Kong SAR and Beijing using facial expression analysis with the AFFDEX algorithm~\cite{AFFDEX}, which was trained on large-scale naturalistic datasets covering diverse demographic groups and is widely used to infer affective emotional responses. All three experiments used bio-evidence collection rather than self-reported surveys, reducing potential bias. This objective methodology allows us to treat these results as ground truth. The findings confirm cross-cultural differences in emotional responses to red tape, indicating that cultural context moderates its emotional impact.

\subsection{Human–AI Alignment Experiment}
\label{subsec:human-ai-align}

\textbf{Culture-aware Persona Construction}
\label{subsec:culture-aware}
To construct agent personas equipped with cultural information, we incorporate demographic characteristics replicated from human experiments, along with family background, profession, personality traits, and historical attitudes toward culturally contingent policies, to enable few-shot conditioning. We employ Hofstede's 6-D model of national culture~\cite{culturefactor_intercultural_management} as the source of country-level cultural values. Specifically, we treat the Hofstede scores as means, assume a standard deviation to account for individual variation, and sample each agent's cultural factor values accordingly. Appendix~\ref{subsec: culture-factor} shows the cultural factor values used for each region. Moreover, we also incorporate language variation in the simulations as part of cultural conditioning to match the human study setup, which was conducted in participants’ regional official languages: German (Germany), Chinese (Mainland China), and English (Hong Kong SAR).

\textbf{Replication of Human Experiments}
Following the human experimental design, we separate agents into control (non-red-tape) and red-tape groups. We then simulate emotional responses under each condition. While human experiments further subdivided red tape manipulation into \textit{Burden} and \textit{Delay} conditions, as mentioned in Section~\ref{subsec: gt}, to capture fine-grained effects, we focus on the primary red tape versus control distinction to examine macro-level cross-cultural differences. We use the identical red-tape scenario descriptions used in the human experiments, which are provided in Appendix~\ref{subsec: scenario-description}. The human-experiment dataset contains emotion-probability scores derived from video-based facial-expression analysis as discussed in Section~\ref{subsec: gt}. We prompt LLMs to output emotion-probability vectors rather than single discrete emotions, since human experimental analyses yield non-zero probabilities across multiple emotions. Our simulations use the same emotions as human experiments, as mentioned in Section~\ref{subsec:human-ai-align}.

\textbf{Model Probing} LLMs exhibit systematic biases, particularly on political and cultural topics~\cite{10.1145/3708319.3733685, chuang-etal-2024-simulating, jiang2024donald, ki-etal-2025-multiple}. To evaluate these biases and assess whether cultural prompting can mitigate them, we design two agent types: default agents and culture-aware agents. Default agents receive red tape scenarios in English without persona or cultural information, establishing a baseline for comparison. Culture-aware agents incorporate the cultural context described above. We apply identical red-tape scenarios to both agent types and compare their emotional responses with human ground truth to determine how much cultural prompting improves models' understanding of culture-specific emotions.

\section{Evaluation}
\label{sec:results}
\subsection{Evaluation Metrics}

The human experiments included 120--150 participants per region. In our simulation setup, we created 200 agents for each region. To reduce noise caused by generation stochasticity, we run 5--10 simulations per model per condition with different random seeds. We evaluate four state-of-the-art general-purpose LLMs: GPT-4o, GPT-5, Gemini-3-Pro, and Qwen3-max-2026-01-23. Due to limited access to the human-experiment datasets, we use the three most salient emotions reported in human results as ground truth for qualitative alignment analysis.

\textit{Top-$K$ Emotion Overlap (Overlap@3).}
To quantify agreement between LLM-predicted emotional responses and human responses in the red tape condition, we adopt the $\mathrm{Overlap@3}$ metric. Specifically, for each culture and condition, we extract the top-$3$ emotions identified in the human experiment, denoted as $H_3$, and the top-$3$ emotions generated by the LLM, denoted as $M_3$. Overlap@3 is defined as the Jaccard similarity between the two sets:
$\mathrm{Overlap@3} = \dfrac{|H_3 \cap M_3|}{|H_3 \cup M_3|}$.
This metric ranges from $0$ to $1$, with higher values indicating stronger agreement between the LLM and human rankings of the most salient emotional responses.

\paragraph{Significance Alignment Score (SAS).}
Since the human baseline from prior work consists of aggregate statistical conclusions rather than raw response distributions, standard distributional metrics such as Jensen--Shannon divergence cannot be computed. We therefore introduce the \emph{Significance Alignment Score} (SAS), denoted by $P_c(R)$, to measure how closely a model's statistical outcome $R$ (Appendix~\ref{subsec:sas-calc}) reproduces the human-observed pattern for cultural group $c$.

Because these patterns differ qualitatively across cultures, the mapping is culture-specific: Hong Kong SAR uses $P_{\mathrm{HK}}(R)=-R$, Germany uses $P_{\mathrm{DE}}(R)=R$, and Mainland China uses $P_{\mathrm{CN}}(R)=-|R-T|$, where $T$ is an empirically derived target (Appendix~\ref{subsec:sas-calc}). Higher SAS always indicates stronger alignment with the corresponding human pattern within that culture. Since these transformations are data-driven rather than theoretically unified, SAS values should be compared only within the same cultural context, not across cultures.



\subsection{Alignment Analysis with Human Results}
Tables~\ref{tab:llm_hk},~\ref{tab:llm_cn}, and~\ref{tab:llm_de} present the metrics for the three countries. \textit{Default} refers to experiments using the baseline agents described in Section~\ref{subsec:human-ai-align}, while \textit{Culture-aware} denotes agents prompted with full cultural context.
\emph{Citizens in Hong Kong SAR appear less sensitive to red tape.} Emotions in Hong Kong SAR show relatively uniform values across conditions, yielding no significant emotion patterns and therefore no Overlap@3 metric. Lower significance levels and higher $SAS$ indicate better alignment with human results. In Table~\ref{tab:llm_hk}, Gemini-3-pro demonstrates the strongest alignment with human significance patterns ($SAS = 0$). However, GPT-5 shows the largest shift in SAS from the Default agent to the Culture-aware agent, suggesting that it absorbs the most cultural information from personas and cultural prompts. \emph{Fear is the distinctive emotion in Mainland China.} The top three emotions observed in the human experiments in Mainland China are \textit{fear}, \textit{surprise}, and \textit{anger}. Differences between the control and red-tape groups are moderate in significance. In Table~\ref{tab:llm_cn}, GPT-5 and Qwen3-max achieve relatively better alignment, though they only reach $\text{Overlap}@3 = 0.2$. Notably, none of the models captured fear among the top three strongest emotions in the simulation experiments, indicating that further research is needed to better simulate fine-grained, human-aligned emotional responses in Eastern cultures. \emph{German citizens are most sensitive to red tape.} The top three emotions observed in the human experiments are \textit{anger}, \textit{frustration}, and \textit{confusion}. Compared with Hong Kong SAR and Mainland China, Germany shows the largest emotional difference between groups. For this culture, higher Overlap@3 and higher $SAS$ indicate better alignment. In Table~\ref{tab:llm_de}, both GPT-4o and GPT-5 achieve perfect accuracy alignment ($\text{Overlap}@3 = 1$), with GPT-5 showing the closest match in significance rate to the human experiments ($SAS = 0.56$). Gemini-3-pro and Qwen3-max also demonstrate improved accuracy compared to their performance in other countries, though they fail to capture the more pronounced emotional changes.

\begin{table}[t]
  \centering

  \small
  \setlength{\tabcolsep}{3.5pt}
  \resizebox{\columnwidth}{!}{%
    \begin{tabular}{lcccc}
      \toprule
      \textbf{Model}
      & \multicolumn{2}{c}{\textbf{Default}}
      & \multicolumn{2}{c}{\textbf{Cultural-aware}} \\
      \cmidrule(lr){2-3} \cmidrule(lr){4-5}
      & \textbf{Overlap@3} & \textbf{SAS}
      & \textbf{Overlap@3} & \textbf{SAS $\uparrow$} \\
      \midrule
      GPT-4o        & - & -0.12 & - &  -0.25 \\

      GPT-5-2025-08-07         & - & -0.50 & - & -0.12 \\

      Gemini-3-pro  & - & -0.11  & - & \textbf{0.00} \\
      Qwen3-max-2026-01-23  & - & -0.13 & - & -0.25 \\
      \bottomrule
    \end{tabular}%
  }
  \caption{Alignment between LLM-simulated and human emotions in Hong Kong SAR. For culture-aware agents, higher SAS means more alignment with human results in significance.}

  \label{tab:llm_hk}
\end{table}

\begin{table}[t]
  \centering
  \small
  \setlength{\tabcolsep}{3.5pt}
  \resizebox{\columnwidth}{!}{%
    \begin{tabular}{lcccc}
      \toprule
      \textbf{Model}
      & \multicolumn{2}{c}{\textbf{Default}}
      & \multicolumn{2}{c}{\textbf{Cultural-aware}} \\
      \cmidrule(lr){2-3} \cmidrule(lr){4-5}
      & \textbf{Overlap@3} & \textbf{SAS}
      & \textbf{Overlap@3 $\uparrow$} & \textbf{SAS} $\uparrow$\\
      \midrule
      GPT-4o        & 0.16 & $-0.0038$ & 0.18 & $\textbf{-0.0038}$ \\
      GPT-5-2025-08-07         & 0.20 & $-0.1338$ & \textbf{0.20} & $-0.1263$ \\
      Gemini-3-pro  & 0.42 & $-0.1263$ & 0.10 & $-0.2463$ \\
      Qwen3-max-2026-01-23 & 0.20 & $-0.1338$ & \textbf{0.20} & $-0.1263$ \\
      \bottomrule
    \end{tabular}%
  }
  \caption{Alignment between LLM-predicted and human emotions in Mainland China. Higher Overlap@3 and higher Significance Alignment Score (SAS) indicate better alignment with human results. (For Mainland China, SAS favors an intermediate significance outcome; see Appendix~\ref{subsec:sas-calc}.) }
  \label{tab:llm_cn}
\end{table}

\begin{table}[t]
  \centering
  \small
  \setlength{\tabcolsep}{3.5pt}
  \resizebox{\columnwidth}{!}{%
    \begin{tabular}{lcccc}
      \toprule
      \textbf{Model}
      & \multicolumn{2}{c}{\textbf{Default}}
      & \multicolumn{2}{c}{\textbf{Cultural-aware}} \\
      \cmidrule(lr){2-3} \cmidrule(lr){4-5}
      & \textbf{Overlap@3} & \textbf{SAS}
      & \textbf{Overlap@3 $\uparrow$} & \textbf{SAS $\uparrow$} \\
      \midrule
      GPT-4o        & 1.00 & 0.22 & \textbf{1.00} & 0.22 \\
      GPT-5-2025-08-07        & 0.93 & 0.56 & \textbf{1.00} & \textbf{0.56} \\
      Gemini-3-pro & 0.25 & 0.00 & 0.50 & 0.00 \\
      Qwen3-max-2026-01-23    & 0.50 & 0.33 & 0.50 & 0.11 \\
      \bottomrule

    \end{tabular}%
  }
  \caption{Alignment between LLM-predicted and human emotions in Germany. Higher Overlap@3 and $SAS$ indicate better alignment.}
  \label{tab:llm_de}
\end{table}



\begin{figure*}[t]
  \includegraphics[width=\linewidth]{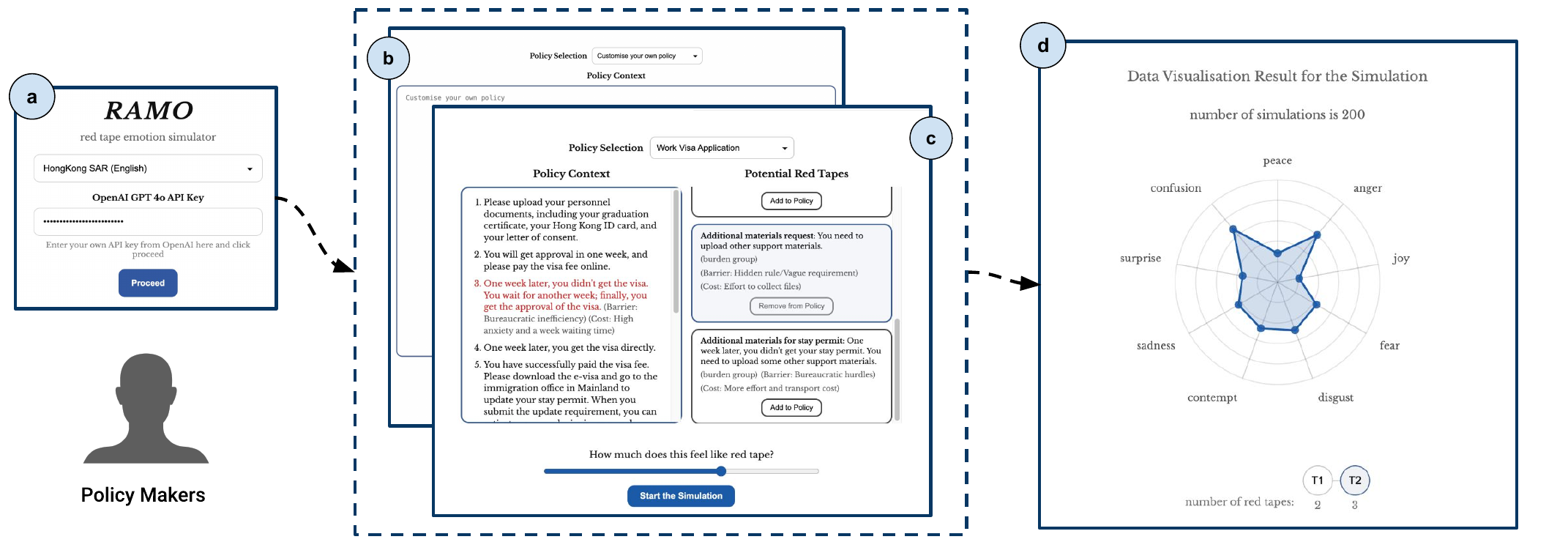}
  \caption{Overview of \textbf{RAMO} interface design, detailed in Section~\ref{sec:ramo_interface}. Major components are enlarged for visibility and labelled with circled letters for reference. The interface is designed for policymakers to simulate potential emotional responses to various red-tape scenarios across cultures. }
  \label{fig:interface_design}
  \vspace{10pt}
\end{figure*}

In conclusion, LLMs demonstrate limited ability to simulate culture-aware, human-aligned emotional responses to red tape. Comparing culture-aware and default agents reveals distinct patterns across models. GPT-5 shows a relatively stronger ability to integrate cultural knowledge through prompting and to mitigate model bias. Gemini-3-pro exhibits a large negative shift under Chinese cultural prompts, indicating poor integration of Eastern cultural contexts. Qwen3-max shows limited effectiveness in incorporating cultural prompts across countries, with the notable exception of Mainland China, consistent with its developmental origins. GPT-4o produces similar results for default and culture-aware agents, indicating that it remains largely influenced by model bias despite cultural prompting. These findings suggest that LLM agents are not yet ready for real-world applications in culturally sensitive domains, such as policymaking, without more sophisticated methods for reducing cultural biases.

\section{RAMO: Red Tape Emotional Response Simulation Interface}
\label{sec:ramo_interface}

We designed an interactive dashboard, named as \textbf{RAMO}, as shown in Figure~\ref{fig:interface_design}, that enables policymakers and stakeholders to execute simulations, author custom policy scripts, and provide self-reported feedback through active interaction.

Upon entering the website, the users will land on the entry page, as shown in Figure~\ref{fig:interface_design} \circnum{a}, featuring the centre title \textbf{RAMO}, and a subtitle, \textit{red tape emotion simulator}. A drop-down menu below the subtitle lets users select their target region. Currently, the system supports Hong Kong SAR, Mainland China, and Germany. To enhance usability and ecological validity for local policy makers, the interface dynamically adjusts its language based on the selection: English for Hong Kong SAR, Simplified Chinese for Mainland China, and German for Germany. This ensures linguistic consistency between the interface and the policy text being analysed. Finally, users must provide a valid OpenAI API key in the designated text field to proceed to the simulation dashboard, as for now we are using GPT-4o as the backend model.

Once the users click the proceed button, they will enter the main dashboard page, divided by a vertical dashed line. The two blocks positioned on either side represent the policy block on the left, as shown in Figure~\ref{fig:interface_design} \circnum{b}, and the result visualisation block on the right, as shown in Figure~\ref{fig:interface_design} \circnum{d}. On the policy block, users can select different policies they want to explore from the drop-down menu at the top. They can explore the predefined policies provided by us as shown in Figure~\ref{fig:interface_design} \circnum{c}. They will be given a list of potential red tapes to select to add to the policy process context box. The steps in the policy will be coloured red if they involve red tape. Users can also customise their policy processes, as shown in Figure~\ref{fig:interface_design} \circnum{b}, but in that case, there will be no listed red-tape items. They will simply be given a text input box to write down the policy. Either way, they would see a progress dragging bar with the instruction, \textit{How much does this feel like red tape?} We intend to collect some self-reported data from this point. If users are satisfied with the policy, they can start running the simulation.

Once the simulation has finished, the result visualisation block, as shown in Figure~\ref{fig:interface_design} \circnum{d}, will render a spider chart, developed using the D3.js library, to illustrate the intensity of the emotions identified in Section~\ref{subsec:human-ai-align}. The interface maintains a session-based history, allowing users to compare results across multiple iterations shown as circled buttons \textit{T1}, \textit{T2}, and so on, as long as the base policy selection remains constant. Each timestamped entry also records the number of red-tape items selected during that run. The number only works for the scenario when users simulate using the predefined policy.

By providing these technical analytical tools, we aim to support policymakers and stakeholders in exploring the micro-level emotional impact of red tape, thereby enabling more informed reflection on policy design and implementation choices. Moreover, RAMO facilitates the collection of human emotional data under dynamically varying red tapes, supporting future improvements in human-model alignment.

\section{Limitations and Future Work}
Due to the limited scope of human experiments, we test and evaluate only one red-tape scenario in a university context, limiting the generalisability of our findings. More human data across diverse red-tape scenarios, particularly in public administration settings, is needed to validate LLM performance. Additionally, while the original studies employed established algorithms trained on large-scale datasets, potential measurement variations across cultural contexts remain a consideration in interpretation. Future work includes model optimisation through enhanced data collection and post-training to provide policymakers with more reliable simulations of emotional responses to red tape.
\clearpage

\bibliographystyle{ACM-Reference-Format}
\bibliography{sample-base}

\clearpage
\onecolumn
\appendix

\section{Supplementary Materials}
\subsection{Culture Factor Table}
\label{subsec: culture-factor}
\begin{figure}[h!]
  \centering
  \includegraphics[width=\columnwidth]{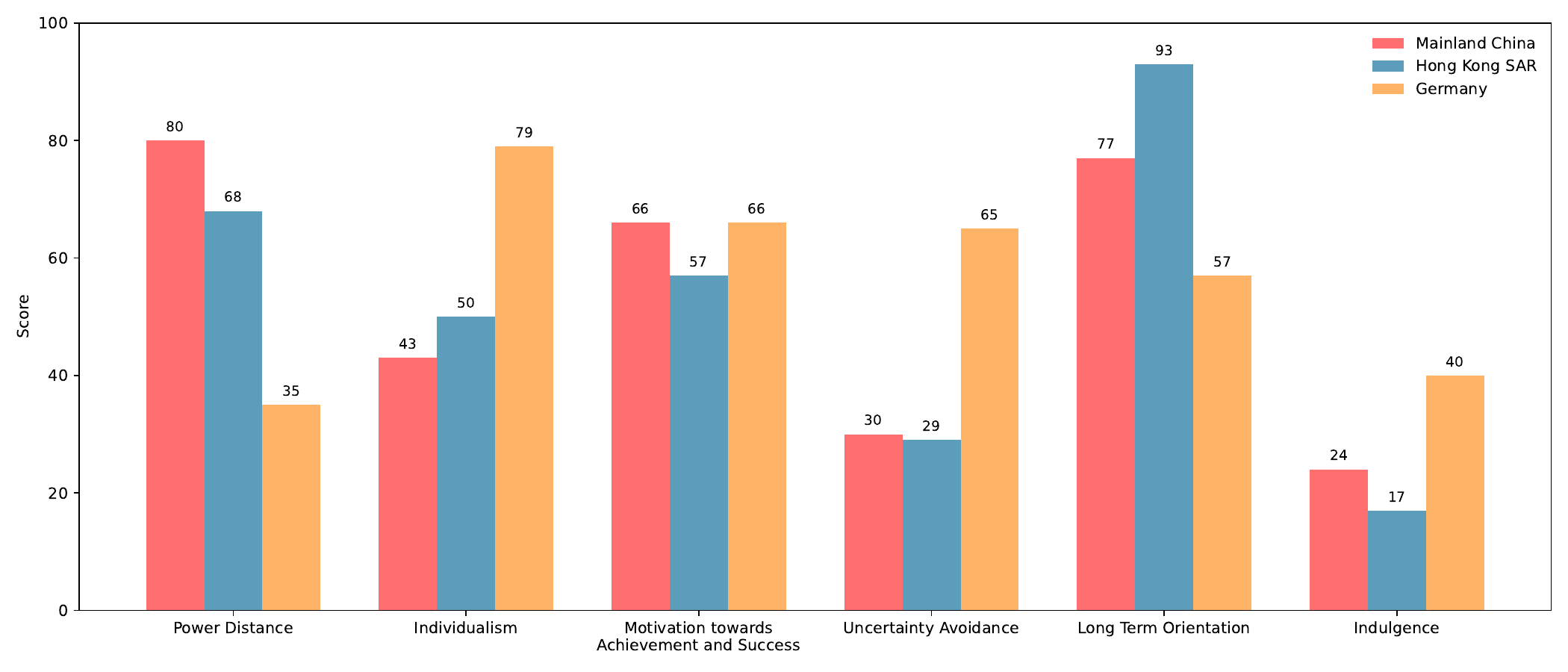}
  \Description{Bar chart (or table-style graphic) comparing Hofstede cultural factor scores for Mainland China, Germany, and Hong Kong SAR across multiple dimensions.}
  \caption{Cultural factor scores of Mainland China, Germany and Hong Kong SAR.}
  \label{fig:cultural_factor}
\end{figure}

\subsection{SAS Calculation}
\label{subsec:sas-calc}

\textit{SigRate$_{95}$ (significance rate at 95\%).}
For each model and region, we measure the proportion of emotions that exhibit a reliable
red-tape effect across repeated paired simulation runs.
Each run $i$ contains both \textit{non-red} and \textit{red-tape} conditions for the same scenario.
Let $\mathcal{E}$ denote the emotion set.
For run $i$, we compute the mean emotion intensity vector
\[
  \bar{\mathbf{y}}_{i}
  =
  \left(\bar{y}_{i,1}, \ldots, \bar{y}_{i,|\mathcal{E}|}\right),
\]ov
by averaging all reactions within the run.

For each emotion $e \in \mathcal{E}$, we define the paired difference
\[
  d_{i,e}
  =
  \bar{y}^{\mathrm{red}}_{i,e}
  -
  \bar{y}^{\mathrm{non\text{-}red}}_{i,e},
\]
and the observed effect
\[
  \Delta_e
  =
  \frac{1}{n}
  \sum_{i=1}^{n} d_{i,e},
\]
where $n$ is the number of paired runs.

We estimate a distribution-free significance threshold using a paired permutation
(sign-flip) test.
For each of 2000 permutations, the sign of $d_{i,e}$ is independently flipped with probability
$0.5$, and $\Delta_e$ is recomputed.
The null distribution of $|\Delta_e|$ is formed by pooling values across all emotions and
permutations.
Let $\tau_{0.95}$ denote the 95th percentile of this pooled null distribution.
We define
\[
  \mathrm{SigRate}_{95}
  =
  \frac{1}{|\mathcal{E}|}
  \sum_{e \in \mathcal{E}}
  \mathbb{I}\!\left[\,|\Delta_e| > \tau_{0.95}\,\right].
\]

\textit{Intermediate target $T$ for Mainland China.}
Let $R$ denote $\mathrm{SigRate}_{95}$.
We define the intermediate target
\[
  T
  =
  \frac{1}{2}
  \left(
    \bar{R}^{\mathrm{def}}_{\mathrm{HK}}
    +
    \bar{R}^{\mathrm{def}}_{\mathrm{DE}}
  \right),
\]
where
\[
  \bar{R}^{\mathrm{def}}_{\mathrm{HK}}
  =
  \frac{1}{|\mathcal{M}|}
  \sum_{m \in \mathcal{M}} R^{\mathrm{def}}_{m,\mathrm{HK}},
  \qquad
  \bar{R}^{\mathrm{def}}_{\mathrm{DE}}
  =
  \frac{1}{|\mathcal{M}|}
  \sum_{m \in \mathcal{M}} R^{\mathrm{def}}_{m,\mathrm{DE}},
\]
and $\mathcal{M}$ is the set of evaluated models.

\subsection{Scenario Description in the Experiments}
\label{subsec: scenario-description}

\begin{quote}
  \small\itshape
  \textbf{Control(non red tape) Group.}
  We have good news for you! You have been selected to receive extra money for taking part in the experiment. Congratulations.
  We will double the amount of money you will get. In other words, you will get 160 HKD for taking part in the experiment.
  You will receive the extra money of 80 HKD straight away. After the experiment, the Research Assistant will present you with the extra money.

  \medskip
  \textbf{Red Tape Group.}
  We have good news for you! You have been selected to receive extra money for taking part in the experiment. Congratulations.
  We will double the amount of money you will get. In other words, you will get 160 HKD for taking part in the experiment.
  You will receive the extra money of 80 HKD at the earliest after 30 days, following university regulations and procedures. It will be transferred to your bank account.
  The Research Assistant will provide you with further instructions after the experiment so we can make the transfer.
  Following university regulations and procedures, you also need to fill in your demographic information again to get the extra money.
\end{quote}

\subsection{Prompt using for simulation}
\label{prompt}
\textbf{China}
\begin{CJK*}{UTF8}{gbsn}
  您是来自中国的BJ001。您是一名 34 岁男性。您拥有学士学位并担任软件工程师。
  您的性格特征包括善于分析、具有社区意识、勤奋、务实。你结婚了。
  大家对你们地区的政策有了一些看法和态度：对近年来的疫情和调控总体持支持态度，认为有利于社会稳定，但也关注民生和中小企业的压力。北京冬奥会展示了城市管理和基础设施能力，让人更期待后续在公共服务上的持续优化。
  就文化因素而言，权力距离得分为 90.51，个人主义得分为 44.77，男子气概得分为 54.19，避免不确定性得分为 13.77，长期取向得分为 80.94，放纵得分为 44.98。
  分数越大表示该维度的趋势越强。

  现在您需要进行公共/政府相关活动：护照申请。政策程序如下：
  1. 请登录“北京公安出境小程序”，注册并预约办理时间，并保存您的预约码。
  2.预约成功。
  3. 请您前往入境大厅，出示您的身份证件，完成指纹采集并提交申请表。
  4.您的申请已受理。护照将在7个工作日内通过EMS快递寄送给您。
  5. 7个工作日后，您顺利收到了EMS快递寄来的护照。
  您对本次活动有何感想？
  想想你这样做是否容易/困难。请遵循您所拥有的文化因素并根据您的文化背景做出反应。
  您对制定这项政策/程序的政府持什么态度？
\end{CJK*}

\textbf{Germany}
Sie sind DE001 aus Deutschland. Sie sind 34 Jahre alt, männlich, haben einen Master-Abschluss und arbeiten als Maschinenbauingenieur.
Zu Ihren Persönlichkeitsmerkmalen zählen Gemeinwohlorientierung, Umweltbewusstsein, Pragmatismus und Pünktlichkeit. Sie sind verheiratet.
Sie haben eine Meinung zur Politik Ihrer Region: Sie unterstützen die Energiewende, wünschen sich aber einen ausgewogenen Übergang, der die Industrie schützt. Sie halten Energiehilfspakete und die Unterstützung der Ukraine für notwendig, bestehen aber auf Haushaltsdisziplin und klaren Zeitplänen.
Im Hinblick auf die Kulturfaktoren erzielten Sie folgende Werte: 41,68 in Machtdistanz, 69,1 in Individualismus, 49,47 in Maskulinität, 54,67 in Unsicherheitsvermeidung, 46,06 in Langzeitorientierung und 48,4 in Genussorientierung.
Ein höherer Wert bedeutet eine stärkere Ausprägung in der jeweiligen Dimension.

Nun müssen Sie an einer öffentlichen/staatlichen Veranstaltung teilnehmen: Visum zur Familienzusammenführung in Deutschland. Das politische Verfahren ist wie folgt:
1. Bitte besuchen Sie die Webseite der „Deutschen Vertretungen in China“, um Ihren Termin zu buchen.
2. Sie haben erfolgreich einen Termin für die nächste Woche gebucht. Bitte laden Sie Ihre Dokumente hoch: Reisepass, Heiratsurkunde, Reisepass des Ehepartners, Meldebescheinigung des Ehepartners in Deutschland, A1-Sprachzeugnis und Krankenversicherung.
3. Beim Termin hat der Sachbearbeiter Ihre Unterlagen entgegengenommen. Ihr Visum wird in 2–4 Monaten fertig sein.
4. Zwei Monate später haben Sie das Visum per EMS erhalten.
Was denkst du über diese Veranstaltung?
Überlegen Sie, ob es für Sie einfach/mühsam ist, das zu tun. Bitte beachten Sie die kulturellen Faktoren, die Sie haben, und reagieren Sie entsprechend Ihrem kulturellen Hintergrund.
Und wie ist Ihre Haltung gegenüber der Regierung, die diese Richtlinie/dieses Verfahren erlassen hat?

\textbf{Hong Kong}
You are HK001 from Hong Kong. You are a 34 years old male. You have a bachelor degree and work as a software engineer.
Your personality traits include analytical, community-minded, diligent, pragmatic. You are married.
You got some opinions and have some attitudes to your region's policy: You generally support the recent epidemic prevention and control measures and housing regulation, believing they contribute to social stability, but are also concerned about the pressure on people's livelihoods and small and medium-sized enterprises. The Beijing Winter Olympics showcased the city's management and infrastructure capabilities, raising expectations for continued optimization of public services.
Culture factor wise, you got a score of 90.51 in power distance, 44.77 in individualism, 54.19 in masculinity, 13.77 in uncertainty avoidance, 80.94 in long term orientation, and 44.98 in indulgence.
Larger score denotes stronger tendency in that dimension.

Now you need to go through public/goverment related event: Beijing Passport Application (red tape). The policy procedure is as follows:
Please log in to the "Beijing Public Security Exit-Entry Mini Program", register and make an appointment. [Click to make an appointment]
<The system shows that all slots are full for the next 30 days.>
<You refresh the page every day. After a week, you finally get an appointment slot.>
Register and make an appointment, and save your appointment code. [Click to make an appointment]
Appointment successful. Please go to the Exit-Entry Hall, show your ID card, complete fingerprint collection and submit the application form. [Click to submit application]
Your application has been accepted. The passport will be sent to you via EMS express within 7 working days.
<After 7 working days, you have not received the passport.>
<The system status shows: Due to "system upgrade and maintenance" and "important meeting security check", your document processing will be delayed.>
<After another 10 days, you finally received the passport sent by EMS express.>

What's your feeling towards this event?
Think about if it easy/troublesome for you to do that. Please follow the cultural factors you have and make the reaction based on your cultural background.
And what's your attitude towards the government who made this policy/procedure.

\end{document}